\documentclass[10pt,twocolumn,letterpaper]{article}

\usepackage{iccv}              

\usepackage{colortbl}

\usepackage{multirow}
\usepackage{booktabs}
\usepackage{tabularx}
\usepackage{arydshln}
\usepackage{array}
\usepackage{xcolor}
\usepackage{makecell} 
\usepackage{pifont}
\usepackage{tcolorbox}
\usepackage{float}
\usepackage{footnote}
\usepackage{footmisc}

\definecolor{pos-green}{RGB}{34,139,34} 
\definecolor{tbl_color}{RGB}{242,242,242}
\definecolor{color_v_freeze}{RGB}{204, 230, 255}
\definecolor{color_v_train}{RGB}{190, 184, 220}

\definecolor{iccvblue}{rgb}{0.21,0.49,0.74}
\usepackage[pagebackref,breaklinks,colorlinks,allcolors=iccvblue]{hyperref}

\def\paperID{6969} 
\def\confName{ICCV}
\def\confYear{2025}

\title{From Holistic to Localized: Local Enhanced Adapters for Efficient Visual Instruction Fine-Tuning}

\author{
Pengkun Jiao\textsuperscript{1,2}, Bin Zhu\textsuperscript{3}, Jingjing Chen\textsuperscript{1,2}\thanks{Corresponding author}, Chong-Wah Ngo\textsuperscript{3}, and Yu-Gang Jiang\textsuperscript{1,2}\\
\textsuperscript{1}Shanghai Key Lab of Intell. Info. Processing, School of CS, Fudan University \\
\textsuperscript{2}Shanghai Collaborative Innovation Center on Intelligent Visual Computing\\
\textsuperscript{3}Singapore Management University\\
{\tt\small pkjiao23@m.fudan.edu.cn}\\
{\tt\small \{binzhu,cwngo\}@smu.edu.sg, \{jingjingchen,ygj\}@fudan.edu.cn}
}

\begin{document}
\maketitle

\begin{abstract}


Efficient Visual Instruction Fine-Tuning (EVIT) seeks to adapt Multimodal Large Language Models (MLLMs) to downstream tasks with minimal computational overhead. However, as task diversity and complexity increase, EVIT faces significant challenges in resolving data conflicts.
To address this limitation, we propose the Dual Low-Rank Adaptation (Dual-LoRA), a holistic-to-local framework that enhances the adapter's capacity to address data conflict through dual structural optimization. Specifically, we utilize two subspaces: a skill space for stable, holistic knowledge retention, and a rank-rectified task space that locally activates the holistic knowledge.
Additionally, we introduce Visual Cue Enhancement (VCE), a multi-level local feature aggregation module designed to enrich the vision-language projection with local details.
Our approach is both memory- and time-efficient, requiring only 1.16$\times$ the inference time of the standard LoRA method (with injection into the query and value projection layers), and just 73\% of the inference time of a 4-expert LoRA-MoE. 
Extensive experiments on various downstream tasks and general MLLM benchmarks validate the effectiveness of our proposed methods. Our project page are publicly available at \url{https://github.com/pengkun-jiao/Dual-LoRA}

\end{abstract}

\begin{figure}[]
    \centering
    \includegraphics[width=1\linewidth]{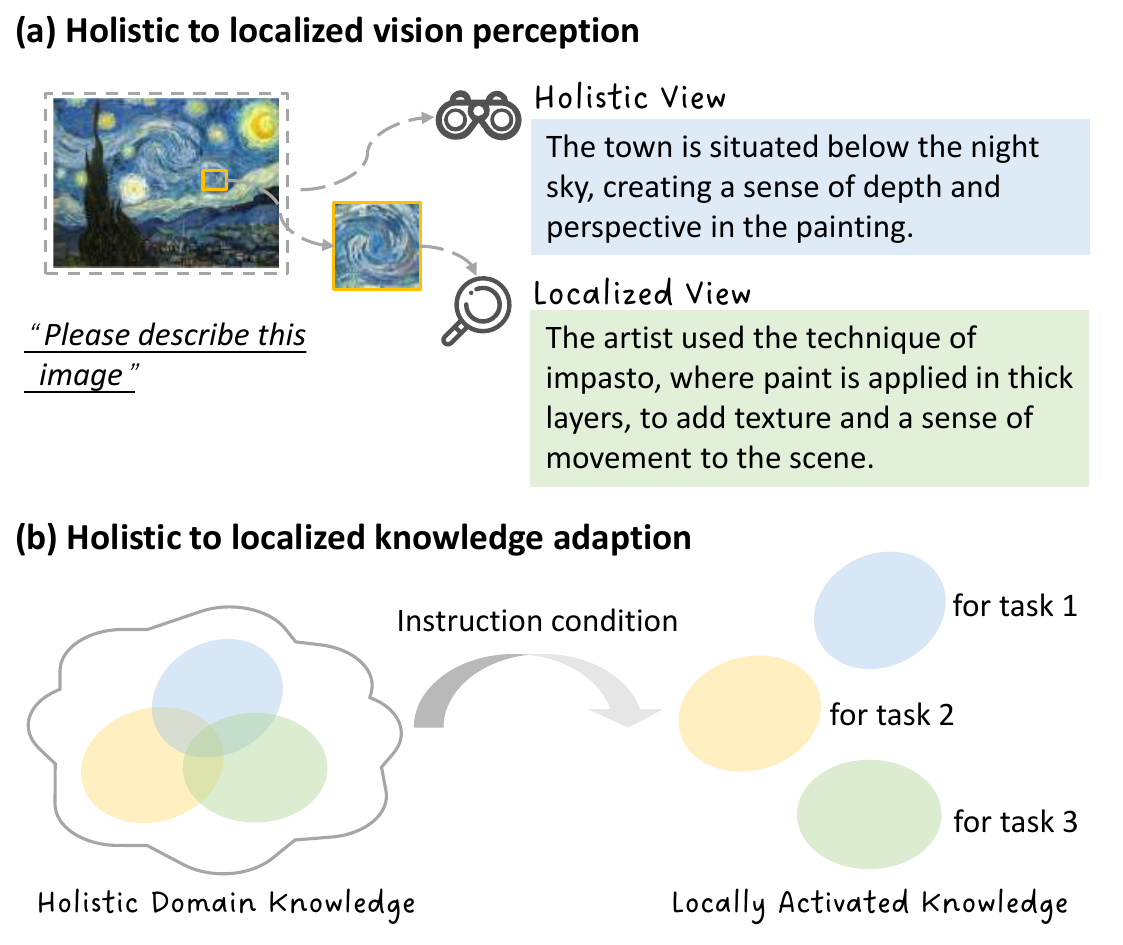}
    \vspace{-.25in}
    \caption{
    The motivation behind the proposed approach. 
    (a) Humans perceive visual signals from both holistic and localized perspectives, considering the overall image structure and local details.
    (b) Humans acquire holistic knowledge, with specific parts (localized) applied to particular tasks. 
    For example, one might first acquire holistic knowledge of cooking and then apply specific knowledge, \eg ingredient recognition, to source ingredients.
    } 
    \vspace{-5pt}
    \label{fig:teaser}
\end{figure}

\section{Introduction}


Multimodal Large Language Models (MLLMs) have demonstrated impressive performances across a wide range of tasks, showcasing versatility and effectiveness in diverse applications~\cite{jiao2025lumen,jiao2025gaseraser,jiao2024inha}. Existing MLLMs~\cite{liu2023llava,bai2023qwen_vl,liu2024llava1.5} typically integrate a pretrained Large Language Models (LLM) \cite{touvron2023llama,chiang2023vicuna} with a pretrained vision encoder \cite{radford2021clip,liu2023dino}, using a vision projector to connect the vision encoder and LLM.

Vision Instruction Tuning~\cite{liu2023llava} enables the practical application of MLLMs in downstream tasks.
However, full fine-tuning MLLMs can be highly resource-intensive. Efficient Vision Instruction Tuning (EVIT) addresses this challenge by tuning only the vision feature projector while integrating a small number of additional trainable parameters (adapters) into the LLM~\cite{ding2023peft}. For example, Low-Rank Adaptation (LoRA)~\cite{hu2021lora} leverages low-rank approximations to replace full fine-tuning of LMM, significantly reducing computational overhead.

\begin{figure*}[t]
    \centering
    \includegraphics[width=1\linewidth]{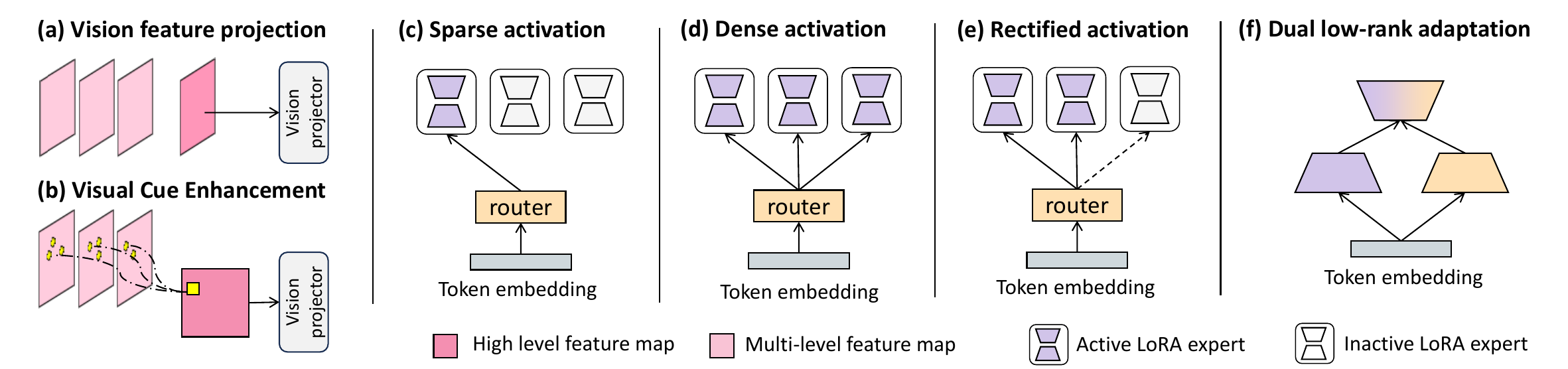}
    \vspace{-0.2in}
    \caption{Comparison between the proposed Dual-LoRA and existing methods.
    (a) Mainstream MLLMs, \eg LLaVA~\cite{liu2023llava} and Qwen-VL~\cite{bai2023qwen_vl}, project the high-level visual feature map. (b) Our Visual Cue Enhancement module enhances high-level features by aggregating local information from multi-level feature maps.
    LoRA-MoE methods mitigate data conflicts by enabling localized responses, i.e., experts activation, tailored to several activation strategies: (c) Sparse Activation\cite{chen2024llava_mole}, where only the top-k experts are activated; (d) Dense Activation\cite{wu2024mixture_of_lora_experts}, where all experts are activated; and Rectified Activation~\cite{jiao2024rode}, where multiple heterogeneous experts are dynamically activated. In contrast, our (f) Dual Low-Rank Adaptation rectifies a holistic knowledge (skill) space with an additional task space, which is fully differentiable, capable of learning any local response, and more structurally efficient.
    } \label{fig:task_skill_space}
\end{figure*}

However, existing EVIT methods face a critical limitation: as downstream tasks grow more diverse and complex, data conflicts in LoRA-based instruction tuning become increasingly pronounced \cite{chen2024llava_mole, huang2024harder_task_need_more_expert}, ultimately impairing performance. An intriguing observation is that when tuning for a variety of food-related tasks, LoRA fine-tuning can result in knowledge-inconsistent answers. For example, some ingredients may be misaligned between food ingredient recognition and recipe generation, as illustrated in Fig.~\ref{fig:quality_result}.
To address this, recent studies suggest integrating LoRA within a Mixture of Experts (LoRA-MoE) paradigm~\cite{chen2024llava_mole,liu2024adamole,wu2024mixture_of_lora_experts}. 
LoRA-MoE adapts to diverse tasks by activating specific experts, which can be viewed as a process of \textit{localized adaptation}, effectively mitigating data conflicts.
However, the LoRA-MoE method requires a complex design to balance activation strategies, trainable parameters, and task complexity~\cite{jiao2024rode, huang2024harder_task_need_more_expert}. Moreover, it incurs significantly higher time consumption compared to vanilla LoRA when multiple LoRA experts are activated.



To address the high time consumption of experts polling in LoRA-MoE, we propose a structurally efficient dual-space adaptation design that transitions from holistic domain knowledge learning to localized, task-specific adaptation. Additionally, we emphasize the role of localized adaptation in visual perception. A high-level overview of our motivation is illustrated in Figure~\ref{fig:teaser}.
This design embodies a cognitively inspired learning paradigm: it first consolidates generalized skill representations and then dynamically routes task-critical knowledge components during inference, mirroring the transition in human cognition from holistic to localized processing.
Specifically, we propose \textbf{Dual Low-Rank Adaptation (Dual-LoRA)}, which strategically decouples adaptation learning into: 1) skill space learning for holistic domain knowledge acquisition, and 2) task space learning for instruction-conditioned task adaptation. The skill space is parameterized using a low-rank matrix optimized stably for cross-task knowledge consolidation, while the task space employs a rank-rectified matrix that dynamically modulates the skill space for localized task-specific adaptation. To enhance local details for vision feature projection, we further propose \textbf{Visual Cue Enhancement (VCE)}, which enhances high-level visual features with local information aggregated from multiple feature maps. VCE remains computationally efficient by focusing on a limited number of local neighboring regions within each feature map.
Figure~\ref{fig:task_skill_space} illustrates the structurally efficient design of our approach.

Our VCE module is only 5.53 MB in size, and Dual-LoRA outperforms LoRA-MoE methods while using the same number of trainable parameters. The full approach requires just 1.16$\times$ the inference time of vanilla LoRA, and only 73\% of the inference time of 4-experts LoRA-MoE methods. Extensive experiments across a range of downstream tasks—including UniFood, ScienceQA, Flickr30k, and general MLLM benchmarks—demonstrate the effectiveness of our proposed methods.
Our main contributions can be summarized as follows: 
\begin{itemize} 
\item We design a time- and memory-efficient holistic-to-localized approach for EVIT. Extensive experiments demonstrate the effectiveness of our proposed methods.
\item We introduce a lightweight \textbf{Visual Cue Enhancement (VCE)} module to enrich fine-grained visual features for efficient visual instruction fine-tuning.
\item We propose \textbf{Dual Low-Rank Adaptation (Dual-LoRA)}, an efficient approach that mitigates data conflicts for instruction fine-tuning.
\end{itemize}



\section{Related Works}

\textbf{Multimodal Large Language Models (MLLMs)} harness the powerful reasoning capabilities of Large Language Models (LLMs)~\cite{touvron2023llama}, driving significant advancements in vision tasks. Early works, such as Flamingo~\cite{alayrac2022flamingo}, resample visual features and integrate them into attention-based adapter layers within the LLM. BLIP-2~\cite{li2023blip2} introduces a Q-Former, which simultaneously handles cross-modal representation learning and generative learning. 
More recently, \textbf{Vision-Language Instruction Tuning} approaches, \eg Instruct-BLIP~\cite{dai2023InstructBLIP}, LLaVA~\cite{liu2023llava}, and Mini-GPT4~\cite{zhu2023minigpt4}, have curated large-scale, high-quality multimodal instruction data to enhance MLLMs’ ability to follow instructions.
However, fully tuning MLLMs is highly resource-intensive due to the vast number of parameters in LLMs, which hinders the application of MLLMs to downstream tasks.



\vspace{0.1in}

\noindent 
\textbf{Low-Rank Adaptation (LoRA)}\cite{hu2021lora} adapts the model by introducing low-rank matrices while keeping the original model parameters unchanged. Specifically, LoRA adds low-rank matrices to certain layers of the model (e.g., attention layers, feed-forward layers, etc.), enabling the model to fine-tune more efficiently for specific tasks. \textbf{Mixture of Experts in LoRA (LoRA-MoE)}\cite{chen2024llava_mole,jiao2024rode,wu2024mixture_of_lora_experts} has gained popularity for addressing data conflicts during downstream task fine-tuning. LoRA-MoE methods can be grouped by the expert activation techniques: sparse activation~\cite{chen2024llava_mole}, where only the top-k experts are activated; dense activation~\cite{wu2024mixture_of_lora_experts}, where all experts are activated and re-weighted by a router; and rectified activation~\cite{jiao2024rode}, where experts are dynamically activated and optimized for sparsity. RoDE~\cite{jiao2024rode} further introduces heterogeneous experts, where different sets of parameters are used for tasks of varying complexity.
Figure~\ref{fig:task_skill_space} compares LoRA-MoE methods with our approach. LoRA-MoE activates multiple LoRA adapters, increasing inference time, whereas our method employs a single Dual-LoRA for greater inference efficiency.

\vspace{0.1in}

\noindent \textbf{Enhancing Vision Feature Representation in MLLMs} primarily involves vision encoder ensembling, resolution enhancement, and multi-level feature map fusion. 
For vision encoder ensembling, MMVP~\cite{tong2024MMVP} utilizes a Mixture of Features (MoF) strategy to integrate image features from CLIP-ViT~\cite{radford2021clip} and DINOv2~\cite{oquab2023dinov2}. Similarly, MouSi~\cite{fan2024mousi} uses an ensemble technique to leverage the strengths of individual vision encoders, introducing a fusion network to unify outputs from different encoders, such as CLIP, DINOv2, and SAM. DeepSeekVL~\cite{lu2024deepseek_vl} adopts a hybrid vision encoder design, encoding images by combining SigLIP-L for low-resolution inputs and SAM-B for high-resolution inputs. LLaVA-UHD~\cite{xu2024llava0uhd} organizes input images for efficient and scalable encoding, utilizing a compression module to further condense image tokens from visual encoders.
Some approaches focus on fusing multi-level feature maps. For example, Cambrian-1~\cite{tong2024cambrian1} introduces the Spatial Vision Aggregator, a dynamic, spatially-aware connector that integrates high-resolution visual features with LLMs while minimizing token count. 
While most of these methods \textbf{require full fine-tuning of the LLM}, our approach, in contrast, introduces a lightweight Visual Cue enhancement module for efficient fine-tuning.


\begin{figure*}[t]
    \centering
    \includegraphics[width=0.99\linewidth]{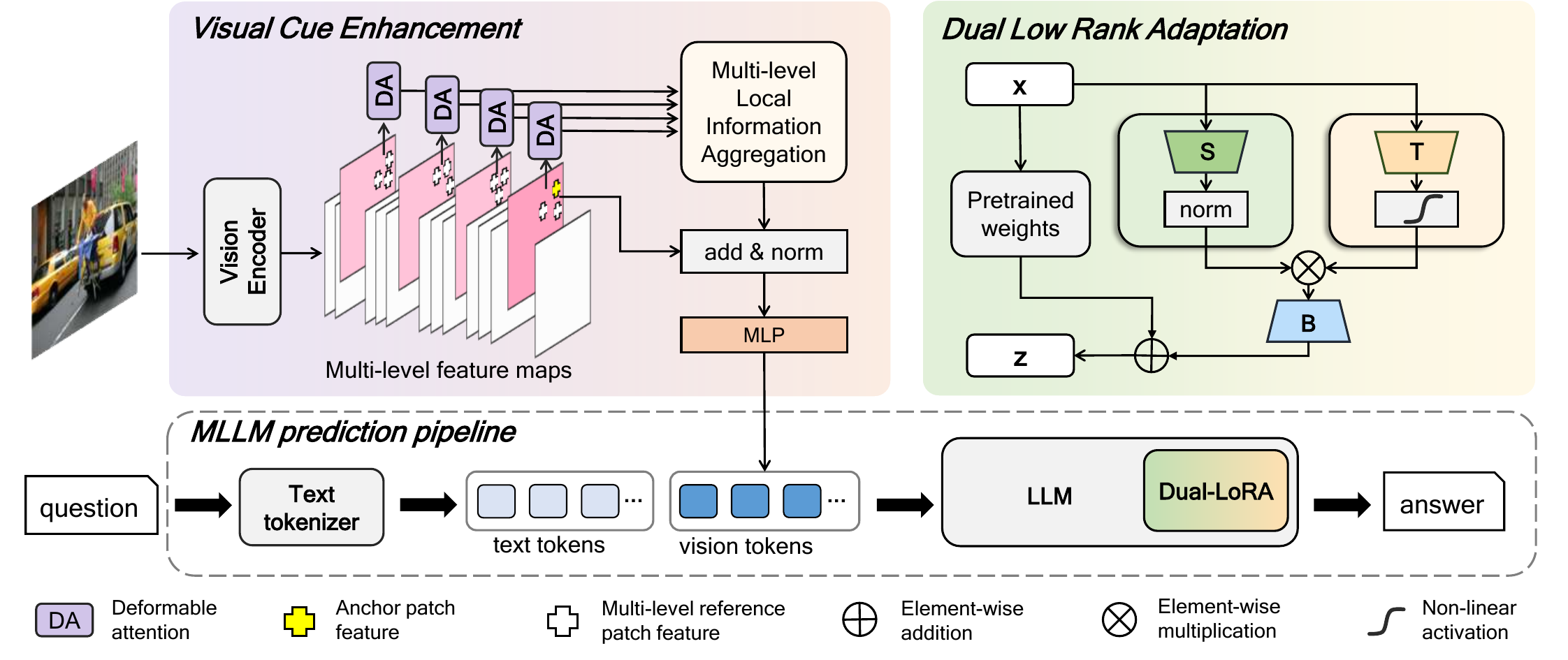}
    \vspace{-.1in}
    \caption{\textbf{Overview of our two key components:} \textbf{Visual Cue Enhancement (VCE)} module enhances high-level anchor features by aggregating local information from multi-level feature maps. The \textbf{Dual Low-Rank Adaptation (Dual-LoRA)} module projects the input feature into two low-rank subspaces: one for stable holistic domain knowledge (skill space) learning and the other for instruction condition (task space) learning. Dual-LoRA modules is integrated into the LLM's query and value projection layers for efficiency.}
    \label{fig:framework}
\end{figure*}

\section{Method}

\subsection{Preliminaries} \label{sec:preliminary}

\noindent \textbf{Multi-Modal Large Language Models (MLLMs).} A standard MLLM (e.g., LLaVA~\cite{liu2023llava}) consists of an image feature encoder \( V(\cdot) \), a vision feature projector \( P(\cdot) \), and a large language model \( G(\cdot) \). Given an image-question-answer (VQA) triplet \((I, T_\text{que}, T_\text{ans})\), \( V \) extracts image features, which are then projected into vision tokens \( t_{\text{vision}} \) through \( P \). These vision tokens are concatenated with question text tokens \( t_{\text{que}} \), and the combined sequence is passed to \( G \) to generate predicted answer tokens \( t_{\text{pred}} \). Instruction tuning aims to optimize the parameters of \( P \) and \( G \) so that \( t_{\text{pred}} \) closely matches the target answer tokens \( t_\text{ans} \):
\begin{align}
\theta_{P}, \theta_{G} &= \arg \min_{\theta_{P}, \theta_{G}} \mathcal{L}(t_{\text{pred}}, t_\text{ans}), \\
t_{\text{pred}} &= G(t_{\text{vision}}, t_{\text{que}}; \theta_{G}), \\
t_{\text{vision}} &= P(V(I); \theta_{P}).
\end{align}

Here, \(\theta_{P}\) and \(\theta_{G}\) represent the trainable parameters of \(P\) and \(G\), respectively, and \(\mathcal{L}(\cdot, \cdot)\) denotes the loss function, typically the cross-entropy loss.

\noindent\textbf{ Low-Rank Adaptation (LoRA).} Due to the large number of parameters in \( G \), optimizing \(\theta_{G}\) can be highly resource-intensive. To address this, LoRA~\cite{hu2021lora} has been proposed as a lightweight adapter injected into the linear layers of LLMs. 
LoRA significantly reduces the number of trainable parameters by decomposing the transformation of the input feature into a low-rank form. The LoRA module is represented by two parameter matrices, \( A \) and \( B \) with the same rank \(r\). In the \(i\)-th injected layer, let \( W_i \) denote the pre-trained weights and \( x_i \) denote the input feature for this layer; then the output after applying the LoRA module can be expressed as:
\begin{align} \label{eq:lora}
z_i &= W_i x + \frac{r}{\alpha } B_i A_i x,
\end{align}
where \( z_i \) is the output feature passed to the next layer, and \(\alpha\) is a scaling factor. For convenience, we omit the indices \( i \) in the following. 
Incorporating LoRA, the trainable parameters of \( F \) are reduced to \(\{\theta_{P}, A, B\}\).





\subsection{Framework Overview}

A visual overview is provided in Fig.\ref{fig:framework}.
The proposed architecture introduces two key components: 1) a Visual Cue Enhancement (VCE) module, detailed in Sec.\ref{sec:vce}, which enhances vision feature projection by aggregating multi-level local visual features, and 2) Dual Low-Rank Adaptation (Dual-LoRA), described in Sec.~\ref{sec:dual_lora}, which employs dual low-rank adaptation pathways to facilitate holistic-to-localized knowledge adaptation.


\subsection{Multi-level Local Visual Cue Enhancement} \label{sec:vce}

Typical MLLMs rely on high-level vision feature maps to align vision and language modalities, \eg LLaVA~\cite{liu2023llava} and Qwen-VL~\cite{bai2023qwen_vl} use the penultimate layer vision feature map of ViT~\cite{radford2021clip}.
While this map primarily captures high-level semantic features and may overlook local visual details.
To address this limitation, we propose a lightweight Visual Cue Enhancement (VCE) module. This module enhances the projected feature map, \ie, the high-level feature \( F^{\text{*}} \), by integrating local information from other layers.

Let $[F_l]_{l=1}^L$ denote multi-level vision feature maps extracted from selected intermediate layers, where $l \in \{1, \ldots, L\}$ indexes the select layers. 
Suppose that the anchor patch feature to be enhanced is $F^{*}(p_q)$, where $p_q \in\mathbb{R}^2 $ is its 2D position.
For each layer $l$, we employ deformable attention~\cite{xia2022deform_attn,zhu2020deformable_detr} to extract local feature $F_{l}'(p_q)$:
\begin{align}
    F_{l}'(p_q) &= \sum_{m=1}^{M} \sum_{k=1}^{K} A_{lmq}(k) \cdot W_{lm} F_l \left( p_q + \Delta p_{lmq}(k) \right), \label{eq:deform_attn} \\
    A_{lmq} &= \mathrm{softmax} \left( \phi_{lm}^{a}(F_l(p_q)) \right), \label{eq:attention_weights} \\
    \Delta p_{lmq} &= \phi_{lm}^{o}(F_l(p_q)), \label{eq:offsets}
\end{align}
\noindent where $m \in \{1, \ldots, M\}$ indexes attention heads, $W_{lm}$ is a learnable projection weight for head $m$ at layer $l$, and $k \in \{1, \ldots, K\}$ indexes reference patches. 
The offset $\Delta p_{lmq} \in \mathbb{R}^{K \times 2}$ represents adaptively learned coordinate offsets around $p_q$, indicate the interest ares,
while $A_{lmq} \in \mathbb{R}^K$ denotes attention weights for input $F_l(p_q)$, where $\phi_{lm}^{o}$ and $\phi_{lm}^{a}$ are linear layers used for learning the offsets $\Delta p_{lmq}$ and attention  $A_{lmq}$, respectively.

A Multi-Level Local Information Aggregation module is then used to consolidate local features across layers:
\begin{equation}
    F'(p_q) = W_o [\mathrm{concat}  (F_{l}'(p_q)) _{l=1}^L], \label{eq:aggregation}
\end{equation}
where \( W_o \) projects the concatenated features into a unified space. 
Finally, the enhanced feature map is obtained through residual fusion and sent to the vision projector to generate enhanced vision tokens:
\begin{equation}
    t_{\mathrm{vision}} = P\left( \mathrm{Norm} \left( F^{*} + \gamma F' \right); \theta_{P} \right), \label{eq:fusion}
\end{equation}
\noindent where \( \gamma \) is a scaling parameter, and \( \mathrm{Norm}(\cdot) \) denotes a LayerNorm~\cite{ba2016layer_norm} layer applied to normalize the residual.

\subsection{Dual Low Rank Adaptation} \label{sec:dual_lora}
In LoRA-MoE, the token embedding is passed through a router, typically a linear layer, followed by an activation strategy that determines which LoRA experts to activate. The activation strategies typically include sparse activation~\cite{chen2024llava_mole}, dense activation~\cite{wu2024mixture_of_lora_experts}, and rectified activation~\cite{jiao2024rode}, each of which has distinct effects across various tasks.

However, efficiently balancing the activation strategy, the granularity of LoRA experts, and task complexity can be challenging. Moreover, the data passing through the router and multiple LoRA experts significantly increases inference time. For instance, a 4-expert LoRA-MoE uses 1.59 times the inference time compared to the non-MoE method (details can be found in Fig.~\ref{fig:time_effeicent}).
Hence, we aim to implement a \textit{unified} approach that achieves the same effect as the LoRA-MoE method but more efficiently. We begin by analyzing the expressiveness of both LoRA-MoE and unified LoRA.

\vspace{0.1in}

\noindent \textbf{Proposition 1:} For a set of LoRAs \( \{ B_k A_k \}_{k=1}^K \), each with rank 1, where \( K \) is the total number of LoRAs, the representation space of a single LoRA with rank \( K \) contains, and is at least as expressive as, the union of the representation spaces of the individual rank-1 LoRAs.

\vspace{0.1in}
\noindent \textbf{Corollary 1:} For a set of LoRAs  \( \{ B_k A_k \}_{k=1}^K \), the combined representation capability does not exceed that of a single LoRA with rank \( \sum \text{rank}(\text{LoRA}_k) \).

\vspace{0.05in}
The proof of \textbf{Proposition 1} and its \textbf{Corollary 1} is provided in Appendix A.

\subsubsection{Learning Holistic Knowledge in One Skill Space}

According to \textbf{Corollary 1}, a single LoRA with equivalent parameters has the same expressiveness as LoRA experts across various ranks. Let \( \mathcal{S} \) denote the adaptation space, which we refer to as the skill space, modeled by the parameter matrix \( S \).

In practice, however, LoRA-MoE often outperforms a single, larger LoRA~\cite{jiao2024rode}. 
\textbf{\textit{What leads to the performance gap, despite they have the same expressiveness?}}
This can be attributed to the differentiated \textit{local response} capability of LoRA-MoE, \ie, expert activation, which provides redundant space to address data conflicts. Two methods ensure the local differences among experts are proposed: 1) variations in experts' weight distributions at a given moment, \eg, RoDE~\cite{jiao2024rode} employs heterogeneous LoRA experts initialized with varying distributions, and 2) differences in routing scores among experts, which can lead to uneven expert activation, \eg, in extreme cases, only a single expert may be utilized~\cite{chen2024llava_mole}.





The observation above suggests that \( \mathcal{S} \) should respond locally (specifically) to the varying tasks. We now consider how to elegantly rectify \( \mathcal{S} \) to achieve a “\textit{local response}”.


\vspace{0.1in}
\noindent \textbf{Corollary 2:} For a LoRA \( BA \) with rank \( r \), the LoRA \( B(A \odot \sigma(T)) \) can be decomposed into any set of LoRA groups \( \{ B_k A_k \}_{k=1}^K \), provided that the constraint \( r \ge \sum_{k=1}^K \text{rank}(B_k A_k) \) holds. Here, \( T \) is a matrix with the same shape as \( A \), and \( \sigma \) is a non-linear activation function. 

The proof of \textbf{Corollary 2} is provided in Appendix A.

\subsubsection{Localized Adaptation by Skill Space Rectification}

According to \textbf{Corollary 2}, additional space can be utilized to map \( \mathcal{S} \) to simulate the “\textit{local response}”, which we refer to as the task space \( \mathcal{T} \). We use a parameter matrix \( T \) to model \( \mathcal{T} \), \( T \) is mapped through a non-linear activation \( \sigma \) to achieve the effect of rank rectification. In this paper, we use the ReLU~\cite{krizhevsky2012cnn} activation function due to its sparsity-inducing properties and ease of optimization~\cite{glorot2011sparseRELU,jiao2024rode}. The rectified space, mapped from the skill and task spaces, can then be expressed as:
\begin{align} \label{eq:dot_multiply_AB}
     D(x) &= \frac{r}{\alpha }  B ( S x \odot \sigma(T x) ),
\end{align}
where \( D \) denotes the Dual-LoRA module. To smooth the distribution of the skill space, and improve stability in subsequent computations, we normalize the skill space using layer normalization~\cite{ba2016layer_norm}, which can be easily implemented thanks to the holistic skill space design. With normalization applied to the skill space, Equation~\ref{eq:dot_multiply_AB} can be revised as:
\begin{align}
    D(x) &= \frac{r}{\alpha }  B ( \text{Norm}(S x) \odot \sigma(T x) ).
\end{align}

Finally, we combine the adaptation feature generated by \( D \) with the feature generated from the pre-trained weight \( W \) to obtain the output feature \( z \):
\begin{align} \label{eq:dual-lora}
    z &= Wx + \frac{r}{\alpha } B ( \text{Norm}(S x) \odot  \sigma(T x) ).
\end{align}

\noindent \textbf{Remark:} Dual-LoRA has a much simpler structure than LoRA-MoE and offers significantly higher time efficiency, an inference time effective comparison is shown in Fig.~\ref{fig:time_effeicent}.

\subsection{Training}
We employ a two-stage fine-tuning framework, consisting of: Stage 1: Vision Projector Pre-training, where only the VCE module and vision feature projector are trainable; and Stage 2: Visual Instruction Fine-tuning, where the VCE module, the vision projector, and the injected Dual-LoRA module are all set to trainable.





\begin{table*}[t]
\resizebox{\textwidth}{!}{
\begin{tabular}{l|cc|cc|cccccccc}
\toprule
\multirow{2}{*}{Method}  & \multicolumn{2}{c}{\textit{Ingredient Recognition}} & \multicolumn{2}{c}{\textit{Recipe Generation}} & \multicolumn{6}{c}{\textit{Nutrition Estimation} (pMAE \(\downarrow\) )}  \\
\cmidrule(lr){2-3}  \cmidrule(lr){4-5} \cmidrule(lr){6-12} 
                               & IoU \(\uparrow\)         & F1 \(\uparrow\)       &  SacreBLEU  \(\uparrow\)      & Rouge-L \(\uparrow\)      & mass    & cal   & fat   &  protein   & carb    & avg    \\
\midrule
vanilla LoRA \cite{hu2021lora}                          &        23.2         & 34.1               & 12.4            & 40.1              &  46.2 & 45.5 & 57.1 & 53.4 & 48.7 & 50.2 \\
\midrule
LoRA-MoE (top-2) \cite{wu2024mixture_of_lora_experts}           &   22.9             &   33.8             &   12.7          &     40.2            & 45.56 & 45.8 & 56.9 & 54.4 & 48.0   & 50.1       \\
\midrule
LoRA-MoE (softmax) \cite{wu2024mixture_of_lora_experts}       &  22.7            &        33.5         &  12.5          &     40.0           &  \underline{45.3}    & \underline{45.5}    & 58.1    &  53.7   &  \textbf{47.5}    &  50.0      \\
\midrule
RoDE \cite{jiao2024rode}                           &  23.6            & 34.6             & 13.8             & 41.4         &  45.8    &  47.6   & 58.5     &  54.4    & 50.4  &  51.3 \\
\midrule
 \textbf{Dual-LoRA}              & \underline{24.2}  & \underline{35.2} & \textbf{14.8}  & \underline{42.1} &  46.1    &  46.2    &  \textbf{56.8} & \underline{52.2} & 48.7 & \underline{49.9} \\
 \textbf{Dual-LoRA + VCE}               & \textbf{24.5}  & \textbf{35.5} &  \underline{14.7}        &  \textbf{42.2} &   \textbf{44.9}    &   \textbf{45.0}    &  \textbf{56.8} &  \textbf{51.2}  &  \textbf{47.5} &  \textbf{49.1} \\
\bottomrule
\end{tabular}
}
\caption{Performance comparison with state-of-the-art Mixture of Experts in LoRA approaches on UniFood dataset. All methods use the same base pretrained MLLM, LLaVA-1.5-7B. The best performance is highlighted in \textbf{bold}, and the second-best is \underline{underlined}.}
\label{tbl:unifood_results}
\vspace{-0.1in}
\end{table*}

\begin{table}[t]
\centering
\resizebox{0.48\textwidth}{!}{%
\begin{tabular}{l|c|cc}
\toprule
\multirow{2}{*}{Method} & \textit{ScienceQA} & \multicolumn{2}{c}{\textit{Flickr30k}} \\
 & Accuracy $\uparrow$ & ScareBLEU $\uparrow$  & Rouge-L $\uparrow$ \\
\midrule
Vanilla LoRA & 70.01 & 27.89 & 66.76 \\
\midrule
LoRA-MoE (top-2) & 76.3 & 28.15 & 66.58 \\
\midrule
LoRA-MoE (softmax) & 77.73 & 28.06 & 67.18 \\
\midrule
RoDE & 78.39 & 28.17 & 66.89 \\
\midrule 
\textbf{Dual-LoRA }& \underline{79.17} & \underline{28.25} & \underline{67.35} \\
\textbf{Dual-LoRA+VCE} & \textbf{80.01} & \textbf{28.71} & \textbf{67.86} \\
\bottomrule
\end{tabular}
}
\caption{Accuracy of single-choice questions on ScienceQA and Image captioning performance on Flickr30k. The best performance is highlighted in \textbf{bold}, and the second-best is \underline{underlined}.}
\label{tbl:result_sciqa_flickr30}
\vspace{-0.1in}
\end{table}

\section{Experiments}

\subsection{Experimental setup}
\textbf{Datasets and evaluation metrics.}
We begin by evaluating our proposed method on downstream tasks using the following datasets and metrics:
1) \textbf{UniFood} \cite{jiao2024rode} is a food-related dataset that includes: ingredient recognition, recipe generation, and nutrition estimation tasks.  For ingredient recognition, we use Intersection over Union (IoU) and F1-score as evaluation metrics. For recipe generation, we assess performance using SacreBLEU~\cite{post2018bleu} and Rouge-L~\cite{lin2004rouge}. For nutrition estimation, we use the percentage of mean average error (pMAE)~\cite{yin2023foodlmm} as the evaluation metric.
2) \textbf{Flickr30k}~\cite{plummer2015flickr30k} is a large image description dataset, with each sample having 5 descriptions. All 5 descriptions are used for training to simulate potential data conflicts, SacreBLEU and Rouge-L metrics are used for evaluation.
3) \textbf{ScienceQA}~\cite{saikh2022scienceqa} is a large-scale multimodal science question-answering benchmark covering diverse subjects. We evaluate it using single-
choice questions, with average accuracy as the metric.
All datasets are trained for one epoch on the training set and evaluated on the test set.

\vspace{0.05in}
\noindent \textbf{Baseline methods.}
We select several Mixtures of LoRA Experts (LoRA-MoE) methods within the LoRA framework: a sparse activation method, LoRA-MoE (top-2)~\cite{wu2024mixture_of_lora_experts}, which activates the top-2 experts; a dense activation method, LoRA-MoE (softmax)~\cite{wu2024mixture_of_lora_experts}, which uses a softmax router to engage all experts; and a rectified activation method, RoDE~\cite{jiao2024rode}, with rectified diverse expert activation. Both LoRA-MoE (top-2) and LoRA-MoE (softmax) are configured with ranks \([16, 16, 16, 16]\), while RoDE is set with varying ranks \([32, 16, 8, 8]\). Additionally, we include a standard LoRA baseline with a rank of 64, and our Dual-LoRA is also set to rank 64.

\vspace{0.05in}
\noindent \textbf{Implementation Details.}
We use the widely adopted open-source model \texttt{LLaVA-1.5-7B}~\cite{liu2023llava} for the base MLLM. 
We conduct our experiments on a setup with 8 RTX 4090 GPUs. The LLM adapters are integrated only within the linear layers of the query and value projectors in the transformer blocks, allowing us to train the model on a single RTX 4090 GPU if necessary.
The detailed base model and optimization configuration can be found in Appendix B.

\vspace{0.05in}
\noindent \textbf{Hyperparameters setting.}
For the VCE module configuration, we select the 2nd, 8th, 14th, and 20th to the last layers of CLIP ViT-L as reference feature maps, with the 2nd layer serving as the anchor feature (see Section~\ref{sec:vce} for details).
We set the scaling parameter \( \gamma \) to 1.0 for all tasks. The VCE module requires only 5.52 MB of memory.  
In the Dual-LoRA setting, we set \( \alpha \) to 2× the rank and the LoRA dropout rate to 0.05.



\begin{table}[]
\centering
\resizebox{0.5\textwidth}{!}{
\begin{tabular}{cc|cccc}
\toprule
\multirow{2}{*}{\textbf{VCE}}  &   \multirow{2}{*}{\textbf{Dual-LoRA}}   & \multicolumn{2}{c}{\textit{Ingredient Recognition}}  & \multicolumn{2}{c}{\textit{Recipe Generation}}       \\
\cmidrule(lr){3-4}  \cmidrule(lr){5-6}
               &                 & IoU                  & F1                   & SacreBLEU                 & Rouge-L              \\
\midrule
  &                    &  23.2  & 34.1   & 12.4   & 40.1   \\
\rowcolor{tbl_color}  &  \checkmark        &   24.2       &   35.2        & \textbf{14.8}     & 42.1           \\
\midrule   
\checkmark       &        &     23.3           &     34.2      &  12.6            &     40.5       \\ 
\rowcolor{tbl_color} \checkmark  &    \checkmark  & \textbf{24.5}  & \textbf{35.5} &  14.7        &  \textbf{42.2} \\
\bottomrule
\end{tabular}
}
\vspace{-0.1in}
\caption{Ablation study of the main components of our proposed method: Visual Cue Enhancement (VCE) and Dual Low-Rank Adaptation (Dual-LoRA). The baseline model vanilla LoRA with LLaVA-1.5-7B is presented in the first row.}
\label{tbl:ablation_component}
\end{table}

\begin{table}[]
\centering
\resizebox{0.5\textwidth}{!}{
\begin{tabular}{cc|cccc}
\toprule
\multirow{2}{*}{\makecell{Skill space \\ @Norm}}  &   \multirow{2}{*}{\makecell{Task space \\ @ Non-linear}}    & \multicolumn{2}{c}{\textit{Ingredient Recognition}}  & \multicolumn{2}{c}{\textit{Recipe Generation}}       \\
\cmidrule(lr){3-4}  \cmidrule(lr){5-6}
      &         & IoU                  & F1                   & SacreBLEU                 & Rouge-L              \\
\midrule
\rowcolor{color_v_freeze}  &       &   21.4          &     32.0         &      13.1               &       41.1          \\
\rowcolor{color_v_freeze}     \checkmark      &          &   22.6          &     33.2         &      14.1               &       41.2            \\
\rowcolor{color_v_freeze}       &     \checkmark       &   21.9               &     32.6         &      13.4               &       41.3          \\
\rowcolor{color_v_freeze}   \checkmark  &    \checkmark       &   23.0       &   33.7        &  14.4              &  41.7               \\
\midrule
\rowcolor{color_v_train}  \checkmark  &  \checkmark        &   \textbf{24.2}       &   \textbf{35.2}        & \textbf{14.8}     & \textbf{42.1}           \\
\bottomrule
\end{tabular}
}
\caption{\textbf{Ablation studies for the design of Dual-LoRA.} ``@Norm'' indicates the use of layer normalization to process the output \(Sx\), while ``@Non-linear'' denotes the use of a non-linear activation function to process the output \(Tx\). The color \textcolor{color_v_freeze}{$\blacksquare$} represents a frozen vision projector, whereas \textcolor{color_v_train}{$\blacksquare$} indicates it is trainable.} 
\label{tbl:ablation_dual_lora}
\vspace{-0.1in}
\end{table}

\subsection{Experiment Results}

\begin{figure}[t]
    \centering
    \includegraphics[width=1\linewidth]{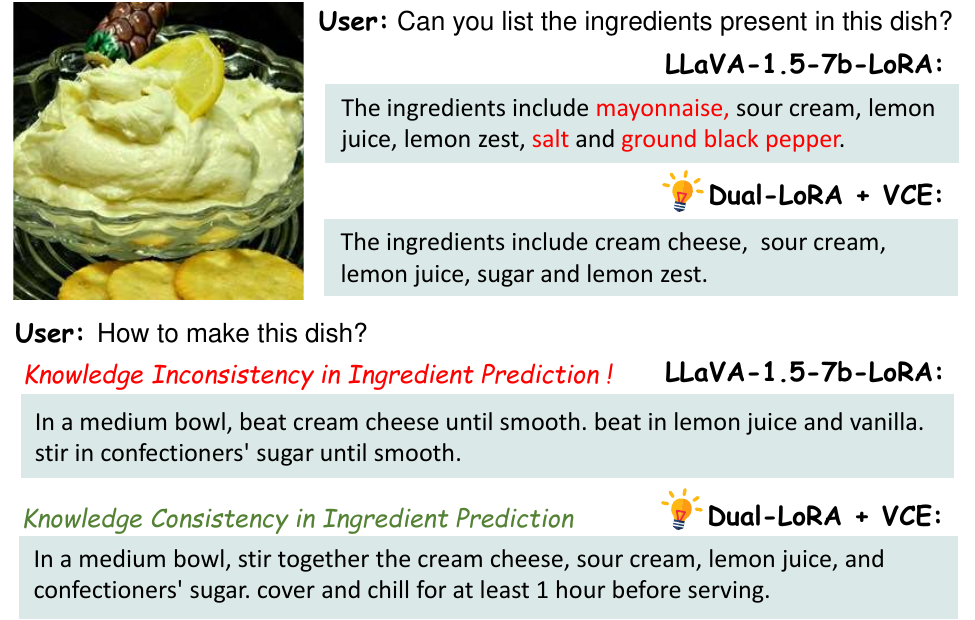}
    \caption{
     Conversations involving two distinct tasks can lead to knowledge-inconsistent responses with LLaVA-1.5-7b-LoRA, due to conflicts arising from multiple complex tasks. In contrast, our proposed method generates more accurate and knowledge-consistent answers.
    }
    \label{fig:quality_result}
    \vspace{-0.1in}
\end{figure}

\textbf{Performance on Multi-task Tuning.}
Considering the data conflicts induced by inconsistencies in optimization goals during multi-task fine-tuning, we first evaluate our approach on the multi-task datasets UniFood and ScienceQA. As shown in Table~\ref{tbl:unifood_results} for the UniFood dataset, our Dual-LoRA method consistently outperforms both vanilla LoRA and LoRA-MoE across all tasks. Similarly, for ScienceQA, as illustrated in Fig~\ref{tbl:result_sciqa_flickr30}, our method surpasses all baseline approaches.
The qualitative results in Fig.~\ref{fig:quality_result} further highlight the effectiveness of our approach, as it consistently generates ingredient lists across different tasks, demonstrating strong alignment with holistic domain knowledge.


\noindent \textbf{Performance on Diverse Annotation Tuning.}
Data conflicts can also arise from diverse annotations within a single task for the same image (\eg varying perspective descriptions of the same image), potentially distorting semantic consistency. We use Flickr30k to test our method and baseline approaches in addressing annotation diversity-induced data conflicts. As shown in Table~\ref{tbl:result_sciqa_flickr30}, our method consistently outperforms the baseline approaches on both SacreBLEU and ROUGE-L scores, further demonstrating the effectiveness of our proposed approach.
We conduct a more in-depth analysis in the following sections.





\subsection{Ablation Study}

\noindent \textbf{Ablation on VCE and Dual-LoRA.}
We conducted an ablation study to evaluate the impact of two key components: the VCE module and the Dual-LoRA module on the UniFood dataset, using the ingredient recognition and recipe generation tasks. The results are presented in Table~\ref{tbl:ablation_component}. This study demonstrates that both the VCE module and the Dual-LoRA module positively influence performance. When both modules are enabled, the best results are achieved for both ingredient recognition and recipe generation tasks.


\vspace{0.05in}
\noindent \textbf{Dual-LoRA design.}
We further investigate the design choices of our Dual-LoRA. Table~\ref{tbl:ablation_dual_lora} presents the results of an ablation study on the normalization of $S$ and the non-linear activation of $T$ (see Section~\ref{sec:dual_lora} for details). This ablation study shows that removing the non-linear activation in the task space or the normalization in the skill space leads to a performance decline. When both are removed, the performance of Dual-LoRA reaches its lowest level.

\begin{figure}[t]
    \centering
    \includegraphics[width=1\linewidth]{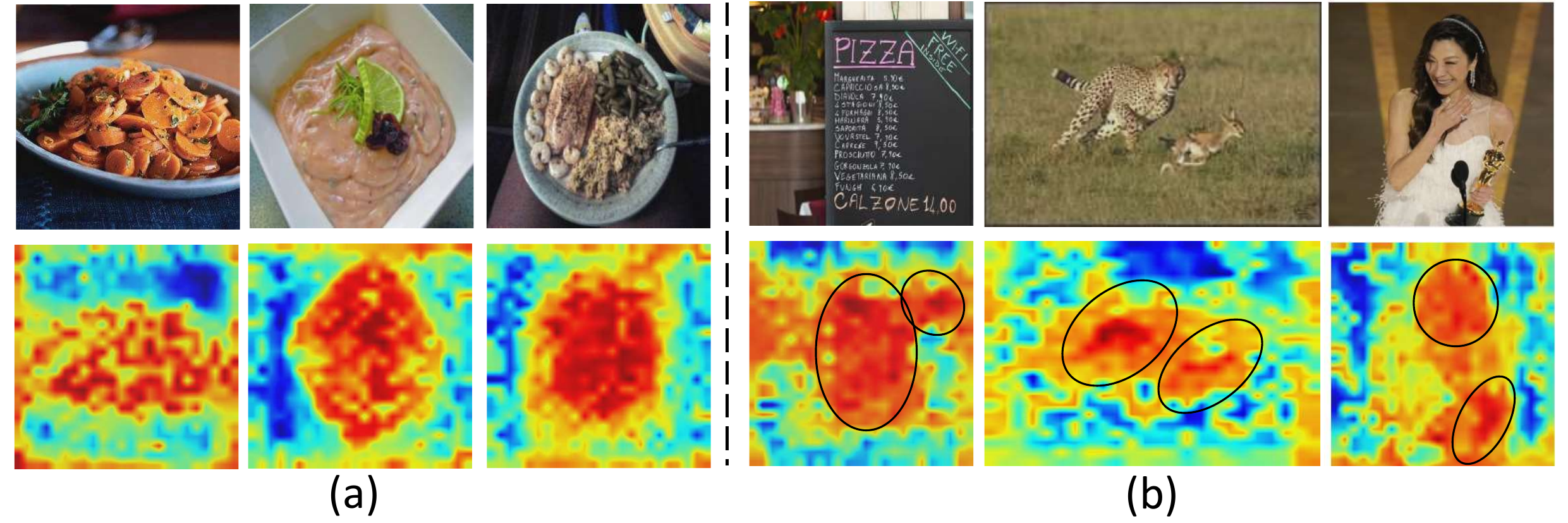}
    \vspace{-0.3in}
    \caption{Feature Map Visualization of Enhanced Visual Cue with VCE:
    (a) Enhanced Visual Cue emphasize key areas in food imagery, such as textures, garnishes, and toppings.
    (b) The cues highlight important details: text for readability, the cheetah-prey interaction, and the woman's face and Oscar trophy on stage.}
    \label{fig:vce_visual_2}
\end{figure}

\begin{figure}[t]
    \centering
    \vspace{-.2in}
    \includegraphics[trim=0cm 0cm 0cm 0cm, clip, width=0.48\textwidth]{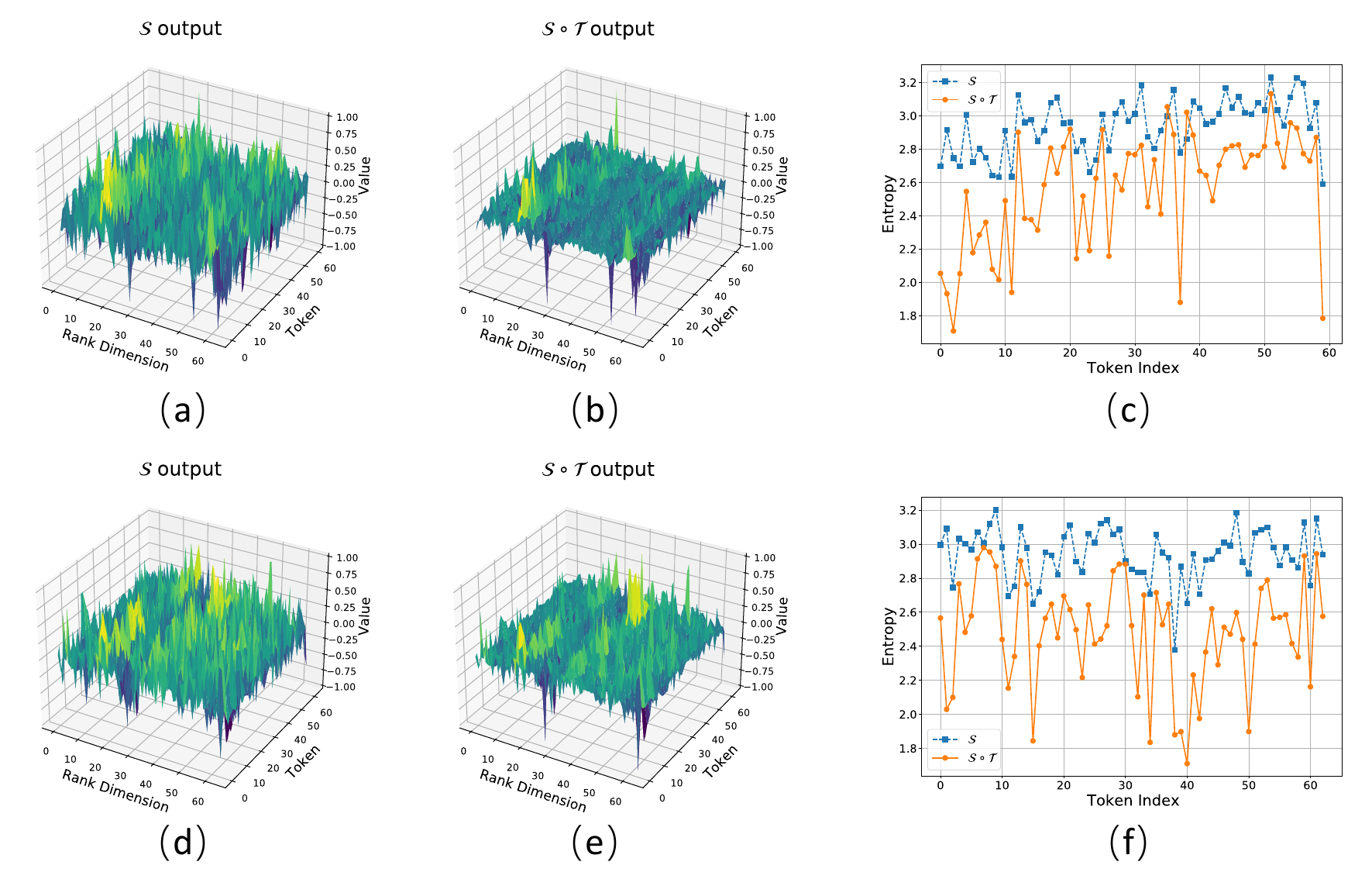}
    \vspace{-.25in}
    \caption{Panels (a), (b), and (c) represent feature visualizations for the recipe generation task where: (a) the distributions of feature outputs for the holistic skill space, (b) the distributions for the rectified skill space, i.e., $\mathcal{S} \odot \mathcal{T}$, and (c) the entropy of the holistic skill space (blue line) and the rectified skill space (orange line). Panels (d), (e), and (f) show the corresponding results for the nutrition estimation task.}
    \label{fig:distribution_dual_lora}
    \vspace{-0.1in}
\end{figure}

\begin{figure*}[t]
    \centering
    \includegraphics[width=1\linewidth]{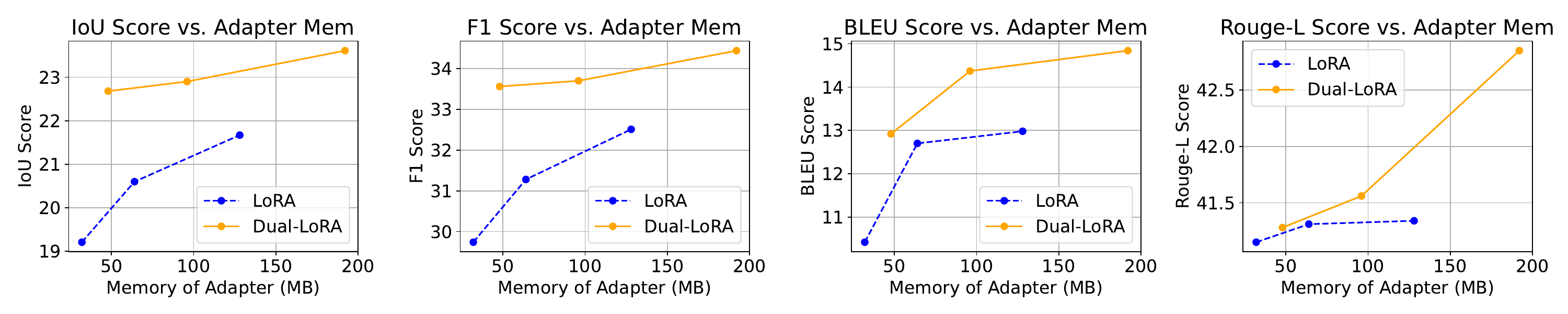}
    \vspace{-0.3in}
    \caption{Performance vs. Trainable LLM Adapter Parameters for Vanilla LoRA and Our Proposed Dual-LoRA on the UniFood Dataset. The adapters are integrated into the query and value projectors of the transformer blocks. To isolate the effect of trainable parameters beyond the LLM adapters, only the adapters are set as trainable.}
    \label{fig:mem_vs_performance}
\end{figure*}

\begin{figure}[t]
    \centering
    \includegraphics[trim=0cm 0cm 0cm 0.3cm, clip, width=0.46\textwidth]{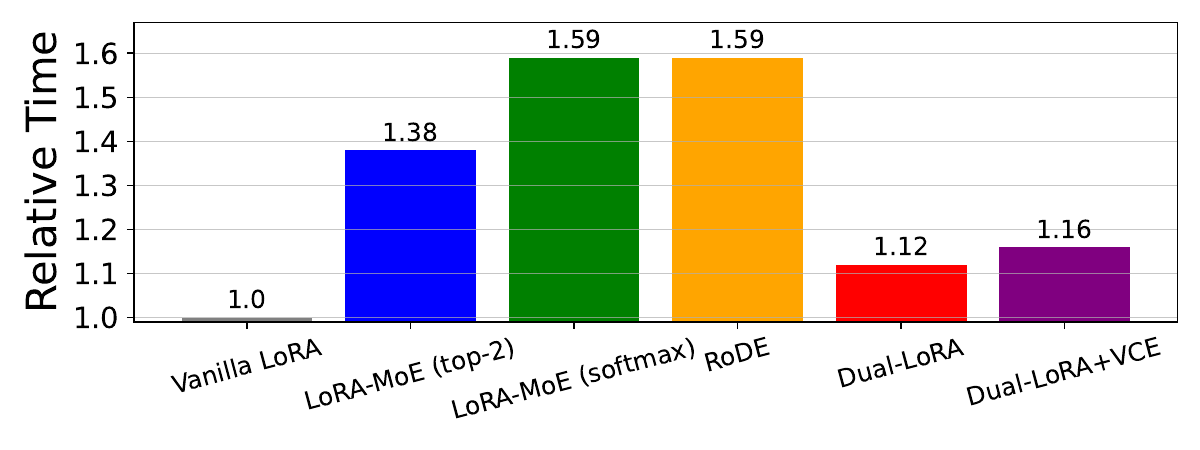}
    \vspace{-0.2in}
    \caption{Relative Inference Time Efficiency of Our Dual-LoRA Method and Several LoRA-MoE Methods. The inference time of vanilla LoRA is used as the baseline (unit 1). The adaptor is injected into the query and value projection layers of the LLM to improve efficiency. Test on the Flickr30k test set.}
\label{fig:time_effeicent}
\end{figure}

\begin{table*}[t]
\centering
\resizebox{.9\textwidth}{!}{
\begin{tabular}{lccc|ccccc}
\toprule
Method & LLM & Vision Encoder & Res. & MMB  & SEED$^\text{I}$ & LLAVA$^\text{W}$ & POPE & MMVet  \\ \hline
InstructBLIP~\cite{dai2023InstructBLIP} & Vicuna-7B  & EVA-CLIP ViT-G & 224 & 36.0 & 60.9 & 58.8 & - & 26.2 \\ 
Shikra~\cite{chen2023shikra} & Vicuna-7B  & CLIP ViT-L/14 & 224 & 58.8  & - & - & -& -  \\ 
LLaVA-1.5~\cite{liu2024llava1.5}  & Vicuna-7B & CLIP ViT-L/14 & 336 & 64.3  & - & 65.4 & 85.9 & 31.1\\ 
LLaVA-1.5-LoRA~\cite{liu2024llava1.5} & Vicuna-7B  & CLIP ViT-L/14 & 336 & \textbf{66.1}   & - & \underline{67.9} & 86.4 & 30.2 \\ 
\midrule
\textbf{Dual-LoRA} & Vicuna-7B & CLIP ViT-L/14 & 336 & 65.1  & 64.5 & 67.4  & \underline{86.8} & \underline{31.5} \\
\textbf{Dual-LoRA+VCE} & Vicuna-7B &  CLIP ViT-L/14 & 336 & \underline{65.6}   & \underline{64.7} & \textbf{68.1}  & \textbf{87.2} & \textbf{32.1} \\ 
 \bottomrule
\end{tabular}
}
\vspace{-0.1in}
\caption{Performance comparison between baseline methods and our proposed method on general benchmarks. The best performance is highlighted in \textbf{bold}, and the second-best in \underline{underlined}.}
\label{tbl:genreral_bench_results}
\vspace{-0.1in}
\end{table*}

\subsection{Qualitative Analysis}

\textbf{Qualitative Results.}
Fig.~\ref{fig:quality_result} presents a comparison between the vanilla LoRA method and our proposed approach. The vanilla LoRA method exhibits knowledge inconsistency, with the ingredient list failing to align with the generated cooking instructions. In contrast, our method produces more accurate and knowledge-consistent results, where the ingredient list aligns with the cooking instructions.
More qualitative results can be found in Appendix D.

\vspace{0.05in}
\noindent \textbf{Qualitative Analysis on VCE.}
Fig.~\ref{fig:vce_visual_2} shows the heatmap of the enhanced Visual Cue, i.e., \( F' \) (see Sec.~\ref{sec:vce} for details). The highlighted areas closely match the regions of interest in the image, demonstrating the effectiveness of our VCE module in enhancing local visual cues.

\vspace{0.05in}
\noindent \textbf{Qualitative Analysis of Dual-LoRA.}
Figure~\ref{fig:distribution_dual_lora} offers a comprehensive visualization of both the skill space and it's rectified space, highlighting several important characteristics: 1) The information entropy of the rectified skill space is lower than that of the original skill space, as shown in panels (c) and (f) of Fig.\ref{fig:distribution_dual_lora}, suggesting that the rectified skill space is more information-determined to the current task. 2) The high-energy frequency bands in the rectified skill space exhibit regional patterns, as evidenced by the varying high values in Fig.\ref{fig:distribution_dual_lora} (b) and (c). These observations demonstrate that the rectified skill space presents a more localized activation of the knowledge, emphasizing task-specific features.

\subsection{Efficiency Analysis}

\noindent \textbf{Memory Usage Efficiency.} Fig.~\ref{fig:mem_vs_performance} presents the results of vanilla LoRA and Dual-LoRA across three rank configurations (32, 64, and 128). As shown, Dual-LoRA consistently outperforms the vanilla LoRA method across all rank settings and tasks. Notably, the performance gap is more pronounced with fewer parameters. 
This can be attributed to the increased visibility of data conflicts when fewer parameters are trainable.
Detailed data on memory usage efficiency comparisons can be found in Appendix E.

\noindent \textbf{Inference Time Efficiency.} Fig.~\ref{fig:time_effeicent} shows the inference time efficiency of our proposed method compared to LoRA-MoE (top-2)~\cite{wu2024mixture_of_lora_experts}, LoRA-MoE (softmax)~\cite{wu2024mixture_of_lora_experts}, and RoDE~\cite{jiao2024rode}, with vanilla LoRA as the baseline (unit 1). The experimental results reveal that LoRA-MoE implementations exhibit significantly prolonged inference durations, with 2-expert and 4-expert configurations requiring 1.38× and 1.59× baseline latency respectively. In contrast, our approach demonstrates exceptional efficiency improvements: the Dual-LoRA configuration achieves only a 1.12$\times$ baseline computational overhead, and when combined with VCE, the complete system maintains high efficiency with merely a 1.16$\times$ increase in baseline latency.


\subsection{Disscusion on General Benchmarks}

We further conduct experiments on general benchmarks for additional analysis.
The results of our proposed method, alongside state-of-the-art MLLMs, are presented in Table~\ref{tbl:genreral_bench_results}, with the experimental setup for these benchmarks detailed in Appendix B. The Dual-LoRA + VCE model consistently ranks among the top two across all benchmark tests and achieves the best performance on LLAVA\textsuperscript{\text{W}}, POPE and MMVet, demonstrating the strong competitiveness of our method on general tasks.


\section{Conclusion}

In this paper, we explore the transition from holistic to localized adaptation for efficient visual instruction fine-tuning. We introduce Dual Low-Rank Adaptation (Dual-LoRA), which enables the model to capture holistic knowledge while being locally activated for specific tasks, thereby mitigating data conflicts. Additionally, we propose Visual Cue Enhancement (VCE), which improves the vision feature projection by incorporating local information from multi-level vision feature maps. Our methods are both time- and memory-efficient, with only a 1.16$\times$ increase in time consumption compared to standard LoRA, yet significantly outperform LoRA in terms of performance with the same parameter settings. Extensive experiments on downstream tasks validate the effectiveness of our approach.

\vspace{0.2in}
\noindent \textbf{Acknowledgements}
This work is supported by the Science and Technology Commission of Shanghai Municipality (Grant No. 24511103100). Additionally, it is partially supported by the Ministry of Education, Singapore, under the Academic Research Fund (AcRF) Tier 2 (Proposal ID: T2EP20222-0046). Any opinions, findings and conclusions or recommendations expressed in this material are those of the authors and do not reflect the views of the Ministry of Education, Singapore.

{
    \small
    \bibliographystyle{ieeenat_fullname}
    \bibliography{main}
}


\end{document}